\documentclass[11pt,a4paper]{article}
\usepackage[hyperref]{eacl2021}
\usepackage{times}
\usepackage{latexsym}
\usepackage{graphicx}
\usepackage{multirow}
\usepackage{textcomp}

\usepackage{microtype}

\aclfinalcopy 

\title{Evaluation of Word Embeddings from Large-Scale French Web Content}

\author{
Hadi Abdine\\
LIX, École Polytechnique\\
hadi.abdine@polytechnique.edu
\And
Christos Xypolopoulos\\
LIX, École Polytechnique\\
christos.xypolopoulos@polytechnique.edu
\AND
Moussa Kamal Eddine\\
LIX, École Polytechnique\\
moussa.kamal-eddine@polytechnique.edu
\And
Michalis Vazirgiannis\\
LIX, École Polytechnique\\
AUEB\\
mvazirg@lix.polytechnique.fr

}

\begin{document}
\maketitle

\begin{abstract}
Distributed word representations are popularly used in many tasks in natural language processing. Adding that pretrained word vectors on huge text corpus achieved high performance in many different NLP tasks. This paper introduces multiple high-quality word vectors for the French language where two of them are trained on massive crawled French data during this study and the others are trained on an already existing French corpus.
We also evaluate the quality of our proposed word vectors and  the existing French word vectors on the French word analogy task. In addition, we do the evaluation on multiple real NLP tasks that shows the important performance enhancement of the pre-trained word vectors compared to the existing and random ones. Finally, we created a demo web application to test and visualize the obtained word embeddings\footnote{\href{http://nlp.polytechnique.fr/}{nlp.polytechnique.fr/}}.
The produced French word embeddings are available to the public, along with the finetuning code on the NLU tasks\footnote{\href{https://github.com/hadi-abdine/WordEmbeddingsEvalFLUE}{github.com/hadi-abdine/WordEmbeddingsEvalFLUE/}}
and the demo code\footnote{\href{https://github.com/hadi-abdine/LinguisticResourcesDemo}{github.com/hadi-abdine/LinguisticResourcesDemo/}}.
\end{abstract}

\section{Introduction}
Word embeddings are a type of word representation that has become very important and widely used in different applications in natural language processing.
For instance, pre-trained word embeddings played an essential role in achieving an impressive performance with deep learning models on challenging natural language understanding problems.
These word vectors are produced using unsupervised neural network models based on the idea that a word's meaning is related to the context where it appears. Thus, similar words in meaning have a similar representations. In result, the quality of word vectors is directly related to the amount and the cleanness of the corpus they were trained on.\\
Many techniques and algorithms have been proposed since 2013 to learn these word distributed representations. Mainly, we focus on Word2Vec \citep{mikolov2013efficient}, GloVe \citep{glove}, and FastText \citep{fasttext} that make use of two approaches: CBoW and SkipGram. These methods were used to learn word vectors using huge corpora. Most of the publicly available word vectors are pretrained on massive English raw text. However, the ones for other languages are few, nonexistent, or trained using a very small corpus that cannot produce good quality word vectors.\\
In this paper, we are interested in creating static French word embeddings with a benchmark for all French static word embeddings. Thus, we do not include contextual word embeddings such as FlaubBERT \citep{le2019flaubert}, CamemBERT \citep{martin-etal-2020-camembert} and BARThez \citep{eddine2021barthez} in the comparison. These attention based pretrained models, even if they have better performance, their cost in terms of both memory and speed is more significant. 
This work presents French word embeddings trained on a large corpus collected/crawled from the French web with more than 1M domains. In addition, we train Word2Vec embedding on other French corpora and resources in order to compare with the ones trained on the collected French corpus. Then, we evaluate these word vectors on the test set created in \citep{grave2018learning} for the word analogy task. Finally, we evaluate the French word embeddings on some tasks from the FLUE benchmark \citep{le2019flaubert} and compare the results with the ones of the existing French FastText \citep{grave2018learning} and Word2Vec \citep{fauconnier_2015} vectors produced using Common Crawl, Wikipedia and FrWac corpus. In the end, we discuss the significance of the word analogy task on word embeddings quality.

\section{Related Work}
Since distributed word representations have been introduced, many pretrained word vectors for many languages are produced. For example, English word vectors were learned using a portion of the google news dataset and published alongside Word2Vec \citep{mikolov2013efficient}. After that, word representations were trained for 157 languages including \textbf{French} using FastText \citep{grave2018learning} to learn word vectors on Common Crawl corpus and Wikipedia. In 2015, \citep{fauconnier_2015} pretrained multiple Word2Vec models for the French language using FrWac corpus \citep{frwacky} and FrWiki dump (French data from Wikipedia). \\
Since Word2Vec first appeared, the evaluation of the word vectors was based on the word analogy task introduced in \citep{mikolov2013efficient}. This task evaluates the quality of word embeddings based on a linear relation between words. Following this idea, a list of French analogy questions to evaluate French word embeddings is introduced in \citep{grave2018learning}.\\
Finally, \citep{le2019flaubert} introduced a general benchmark to evaluate French NLP systems named FLUE containing multiple and diverse French tasks to evaluate French natural understanding models. \\
Our main objective is to evaluate the quality of static word embeddings for the French language and provide new ones trained on a large, diverse dataset that perform well on word analogy tasks, as well on language understanding tasks. Mentioning that, there are many papers showing that pretrained attention based models like BERT \citep{DBLP:journals/corr/abs-1810-04805} and ELMo \citep{elmo} out-perform static word embeddings. However, our intuition is that static word embeddings are still relatively crucial for low resources and small applications where using expensive models that require huge loads of GPU resources is not applicable.

\section{Crawling The French Web}\label{corpus}

\hspace{0.5cm}\textbf{Seeds}
Our initial seeds were selected by finding the most popular web pages under the top level domain ".fr".
Our intuition is that web pages constitute the core of the french web graph.
To achieve this, we used a publicly available list\footnote{\href{https://domaintyper.com/top-websites/most-popular-websites-with-fr-domain}{Most popular websites with fr domain}}.
The frontier was then updated with the new links that were discovered during crawling.

\textbf{Crawler}
For crawling we chose Heritrix\footnote{\href{https://github.com/internetarchive/heritrix3}{https://github.com/internetarchive/heritrix3}}, an open-source web-scale crawler that is supported by the Internet Archive.
Our setup constitutes of a single node with 25 threads crawling for a period of 1.5 months.

\textbf{Output}
The generated output follows the WARC file format, with files split to 1GB each.
This format has been traditionally used to store "web crawls" as sequences of content blocks harvested from the World Wide Web.
The content blocks in a WARC file may contain resources in any format; examples include the binary images or audiovisual files that may be embedded or linked to HTML pages.

\textbf{Extraction}
To extract the embedded text we used a warc-extractor tool.\footnote{\href{https://github.com/alexeygrigorev/warc-extractor}{https://github.com/alexeygrigorev/warc-extractor}}
This tool is used in order to parse the records with the WARC and then parse the HTML pages.
During this step we also integrated the FastText language detection module\footnote{\href{https://github.com/vinhkhuc/JFastText}{https://github.com/vinhkhuc/JFastText}} that allows us to filter the HTML text by language. 

\textbf{Deduplication}
To eliminate redundant data in the corpus, we used Isaac Whitfield's deduplication tool \footnote{\href{https://github.com/whitfin/runiq}{https://github.com/whitfin/runiq}}, the same used on Common Crawl corpus \cite{ortiz2019} based on a very fast and collisions resistant hash algorithm.

\textbf{Ethics}
Heritrix is designed to respect the robots.txt (file written by the website owners to give instructions about their site to web crawlers), exclusion directives and META nofollow tags.

\begin{table*}[ht]
            \centering
            \begin{tabular}{@{}cccccc@{}}
            \hline
            Embeddings  & Vocab & Tool & Method & Corpus  & Dimension\\
            \hline
            fr\_w2v\_web\_w5    & 0.8M  & Word2Vec & cbow & fr\_web  & 300\\
            fr\_w2v\_web\_w20   & 4.4M & Word2Vec & cbow & fr\_web  & 300\\
            fr\_w2v\_fl\_w5 & 1M & Word2Vec & cbow & flaubert\_corpus  & 300\\
            fr\_w2v\_fl\_w20  & 6M  & Word2Vec & cbow & flaubert\_corpus  & 300\\
            \hline
            cc.fr.300\citep{grave2018learning}   & 2M & FastText & skip-gram & CC+wikipedia  & 300\\
            frWac\_200\_cbow\citep{fauconnier_2015} & 3.6M & Word2Vec & cbow & frWac  & 200 \\
            frWac\_500\_cbow\citep{fauconnier_2015} & 1M & Word2Vec & cbow & frWac  & 500 \\
            frWac\_700\_cbow\citep{fauconnier_2015} & 184K & Word2Vec & skip-gram & frWac & 700 \\
            frWiki\_1000\_cbow\citep{fauconnier_2015} & 66K & Word2Vec & cbow & wikipedia dump & 1000 \\
            \hline
            \end{tabular}
            \caption{Summary of the models used in our experiments}
            \label{tab:embeddings}
\end{table*}
            
\section{Word Embeddings Training}
In order to learn the French word embeddings, we used Gensim's Word2Vec to produce four models of CBoW (Continuous Bag of Words) French word vectors.\\ Two of these models were trained on a 33GB shuffled portion of the French corpus used to train FlauBERT \citep{le2019flaubert}. This corpus, consists of 24 sub-corpora collected from different sources with diverse subjects and writing styles. The first model is noted as fr\_w2v\_fl\_w5 trained using Word2Vec CBoW with a window size of 5 and min\_count of 60, in other words, we only consider training an embedding for a word, only if it appears at least 60 times in the corpus. The second one is fr\_w2v\_fl\_w20 trained using Word2Vec CBoW with a window size of 20 and min\_count of 5.\\ The other two models were trained on the 33GB deduplicated French corpus collected from the web (section \ref{corpus}). The first is noted as fr\_w2v\_web\_w5 trained on Word2Vec CBoW with a window size of 5 and min\_count of 60. The second is fr\_w2v\_web\_w20 trained on Word2Vec CBoW with a window size of 20 and min\_count of 5. All the models examined in this paper have an embedding size of 300. Table \ref{tab:embeddings} contains the details of each word embeddings used in this paper.

\section{Word Analogy Evaluation}
In this work, the first evaluation method for the word vectors is the analogy task \citep{mikolov2013efficient}. In this task, given three words, A, B and C, we can predict a word D such that the relation between A and B is the same between C and D with an assumption that a linear relation between word pairs indicates the quality of word embeddings. For example, if the relation between word representations of \textbf{king} and \textbf{man} is similar to the relation between the representations of \textbf{queen} and \textbf{woman}, this will imply good word embeddings. For instance, in this evaluation we use the word vectors $x_A$, $x_B$ and $x_C$ of three words A, B and C to compute the vector $x_B-x_A+x_C$, and the closest vector in the dictionary to the resulted vector will be considered the vector of the word D.\\ The performance of word vectors is finally computed using the average accuracy over the evaluation set.\\
To evaluate our French word embeddings presented in section \ref{corpus}, we use the French analogy dataset created in \citep{grave2018learning}. This dataset contains 31 688 questions. We also compare the results of our word vectors with the one of the French FastText (cc.fr.300), also produced in \citep{grave2018learning} and the French wacky word vectors.

    \begin{table}[ht]
    \centering
    \begin{tabular}{ll}
    \hline
    Embeddings  & Accuracy\\
    \hline
    fr\_w2v\_web\_w5       & 41.95   \\
    fr\_w2v\_web\_w20    & 52.50 \\
    fr\_w2v\_fl\_w5  & 43.02    \\
    fr\_w2v\_fl\_w20   & 45.88   \\
    \hline
    cc.fr.300   & 63.91  \\
    frWac\_500\_cbow & \textbf{67.98}\\
    frWac\_200\_cbow & 54.45\\
    frWac\_700\_sg & 55.52\\
    frWiki\_1000\_cbow & 0.87\\
    \hline
    \end{tabular}
    \caption{Accuracy on the analogy task}
    \label{tab:analogy}
    \end{table}
    
\textbf{Results}    Table \ref{tab:analogy} shows the accuracy on the analogy task for all the models. The results indicate that the FastText word vectors and the French wacky word embeddings are by far better than our CBoW Word2Vec word vectors in the analogy task. However, the authors in \citep{inbook} proved that the property of geometrical relation with respect to analogies does not generally hold and is likely to be incidental rather than systematic. In conclusion, we decided to evaluate the word vectors on natural language understanding (NLU) tasks such as text classification, paraphrasing and language inference that are presented in \citep{le2019flaubert}.\\

\begin{figure*}
  \includegraphics[width=\textwidth,height=2.5cm]{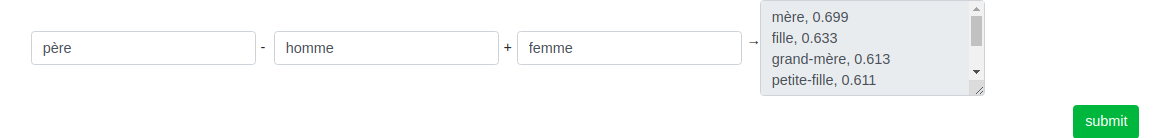}
  \caption{Example of word analogy using the analogy tool in our web app (with fr\_w2v\_web\_w20).}
  \label{analogyfig}
\end{figure*}
\begin{figure*}
  \includegraphics[width=\textwidth]{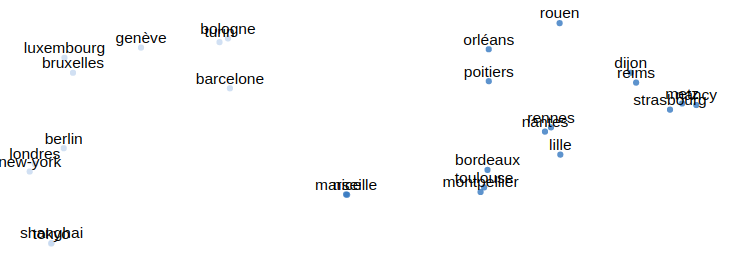}
  \caption{2 of 8 clusters of the query: "Top 200 closest words to \textbf{paris}" using the visualization tool in our web app (with fr\_w2v\_web\_w20)}
  \label{vis}
\end{figure*}
\textbf{Demo and web app}  To visualize and examine our word embeddings, we created a web app that contains the following tools:
\begin{enumerate}
    \item Word analogy examiner: it takes as input three words, A, B and C, and it computes the closest ten vectors to $x_B-x_A+x_C$, and displays their corresponding words as we can see in figure \ref{analogyfig}.
    \item A cosine similarity calculator: it takes two words and computes the cosine similarity between the corresponding vectors.
    \item A similar words tool: find the top ten similar words to an input word.
    \item A visualization tool: using T-SNE and k-means, this tool displays in a 2-D vector space the top \textbf{n} closest words to the word \textbf{w} distributed in \textbf{k} clusters, where n, w and k are chosen by the user. For example, in figure \ref{vis} we see a part of the graph for n=200, w=paris and k=8.
\end{enumerate}

\section{Evaluation On FLUE Benchmark Tasks}
    
   FLUE is a French language understanding evaluation benchmark \citep{le2019flaubert}, created to evaluate the performance of French NLP models. It contains many datasets that vary in subjects, difficulty level, size, and degree of formality. This section presents the results of finetuning of our different Word2Vec embeddings and compares them with the finetuning results of the FastText and the French wacky Word2Vec embeddings on some FLUE benchmark tasks. We also include a non-pretrained word embedding for comparison.
   
        \subsection{Text Classification}
        
        \hspace{0.5cm}\textbf{Dataset description}   In this task, the used dataset is the French portion of CLS (Cross Lingual Sentiment) \citep{prettenhofer-stein-2010-cross}. It consists of Amazon reviews divided into three subsets: books, DVD, and music.
        Following \citep{le2019flaubert}, each subset is binarized by considering all reviews above three stars as positive and the ones below three stars as negative, while eliminating the ones with three stars. In addition, each subset contains a balanced train and test set that has around 1000 positive and 1000 negative reviews each. Finally, a validation set is formed for each subset by taking a random split of 20\% of the training data.\\
        
        \textbf{Model description}   The model uses word embedding layer that varies in initial weights and in dimension according to the evaluated word embedding. This layer is followed by a two-layer 1500D (per direction) Bi-LSTM. Finally, we use the classification head used in \citep{DBLP:journals/corr/abs-1810-04805} that is composed of the following layers: a dropout layer with a dropout rate of 0.1, a linear layer, a hyperbolic tangent activation layer, a second dropout layer with the same dropout rate and a second linear layer with output dimension of two (the same as the number of classes).\\
        We train the model for 30 epochs for all the different word embeddings while performing a grid search over three different learning rates: 5e-5, 2e-5 and 5e-6. The best-performing model on the validation set is chosen for the evaluation on the test set.
        
            \begin{table}[ht]
            \centering
            \begin{tabular}{llll}
            \hline
            Embeddings  & Books & Music & DVD \\
            \hline
            w2v\_0       & 67.20  & 67.20 & 62.15    \\
            \hline
            fr\_w2v\_web\_w5       & 79.38 & 79.49 & 80.29    \\
            fr\_w2v\_web\_w20    & 81.55  & 79.75  & 80.75  \\
            fr\_w2v\_fl\_w5  & 82.16  & 81.00 & 79.64    \\
            fr\_w2v\_fl\_w20   & \textbf{82.38}  & \textbf{82.58}  &  \textbf{82.43}    \\
            \hline
            cc.fr.300   & 75.30 & 74.35 & 72.43   \\
            frWac\_500\_cbow & 81.30 & 80.80 & 79.59\\
            frWac\_200\_cbow & 79.65 & 75.40 & 80.75\\
            frWac\_700\_sg & 77.55 & 75.20 & 78.43\\
            frWiki\_1000\_cbow & 66.30 & 70.05 & 60.94 \\
            \hline
            \end{tabular}
            \caption{Accuracy on the French CLS}
            \label{tab:cls}
            \end{table}
            
        \textbf{Results}   Table \ref{tab:cls} presents the accuracy on the test set for each word embeddings. The results demonstrate the importance of pretrained models. As we see, all pre-trained word embeddings surpass by far the non pretrained embeddings with a difference that can goes to more than 20\% in accuracy. In addition, we see clearly the contradiction in results between the analogy task and the sentiment analysis.
            
        \subsection{Paraphrasing}\label{pawsx}
        \hspace{0.5cm}\textbf{Dataset description}   This task uses the French portion of PAWS-X \citep{Yang_2019} which extends the cross-lingual adversarial dataset for paraphrase identification. This task aims to identify if there is a semantic relation between a pair of sentences or not. The dataset is obtained from translating collected pairs of sentences in English from Wikipedia and Quora with a high lexical overlap ratio and judged by humans. The used dataset contains 49 401 training samples, 1 992 validation samples and 1 985 test samples. 
        
        \textbf{Model description}   To finetune the word vectors on PAWS-X, we use Enhanced Sequential Inference Model (ESIM) \citep{chen2016enhanced}. This model is formed from three essential components: input encoding, local inference modeling and inference composition.\\
        After the word embedding layer, the input encoding layer is composed of 1-layer 1500D BiLSTM that encodes the local inference information of sentences A and B with a length of respectively $l_a$ and $l_b$ to give the output $\overline{A}$ and $\overline{B}$.\\
        The following component is the local inference modeling where:
        \begin{enumerate}
            \item We compute the similarity matrix E between the two sentences representing the attention weights, such that $e_{ij}=\overline{a}_{i}^{T}\overline{b}_{j}$.
            
            \item We compute the local relevance between the two sentences where each word will be represented as a weighted summation of the relevant information to this word. For instance, $\tilde{a_{i}}=\sum_{j=1}^{l_{b}}softmax(e_{ij})\overline{b_{j}}$ and $\tilde{b_{j}}=\sum_{i=1}^{l_{a}}softmax(e_{ij})\overline{a_{i}}$.
            
            \item We find the enhancement of local inference information by a concatenation in a way that the authors expect would help to enhance the local inference. For instance, the resulted vectors will be $m_{a}=[\overline{a}; \tilde{a}; \overline{a} - \tilde{a}; \overline{a} \odot \tilde{a}]$ and $m_{b}=[\overline{b}; \tilde{b}; \overline{b} - \tilde{b}; \overline{b} \odot \tilde{b}]$.
        \end{enumerate}
        The fourth layer is a projection layer that is formed from a linear layer with a hidden dimension of 1500D, ReLU activation and a dropout layer with a rate of 0.1\\
        Finally, the inference composition layer is composed of 1-layer 1500D BiLSTM layer that gives the vectors $V_{a}$ and $V_{b}$ as output. We used the same classification head as in the text classification task. Where the input to the classification head is the concatenated vector of average pooling and max pooling of $V_{a}$ and $V_{b}$.
        
            \begin{table}[ht]
            \centering
            \begin{tabular}{ll}
            \hline
            Embeddings  & Accuracy\\
            \hline
            w2v\_0       & 73.49    \\
            fr\_w2v\_web\_w5       & 77.12    \\
            fr\_w2v\_web\_w20    & \textbf{78.02} \\
            fr\_w2v\_fl\_w5  &75.2    \\
            fr\_w2v\_fl\_w20   &74.24    \\
            \hline
            cc.fr.300   & 62.80  \\
            frWac\_500\_cbow & 71.47\\
            frWac\_200\_cbow & 74.80\\
            frWac\_700\_sg &  67.39\\
            frWiki\_1000\_cbow & 69.76\\
            \hline
            \end{tabular}
            \caption{Accuracy on the French PAWS-X using ESIM}
            \label{tab:pawsx}
            \end{table}
            
        \textbf{Results}   The final accuracy of each word vectors is reported in Table \ref{tab:pawsx}. Again, one can see that not only there are opposed results between NLU tasks and the analogy task, but also that random word vectors outperform the word vectors that achieved the second-highest accuracy in the analogy task.  
        
        \subsection{Natural Language Inference}
        
         \hspace{0.5cm}\textbf{Dataset and model description}    The French natural language inference task in FLUE uses the French portion of XNLI dataset \citep{Conneau_2018}. It contains NLI data for 15 languages. Each pair-sentences in the test and validation sets are annotated manually by humans with three classes of inference: entailment, neutral, and no entailment (contradiction). Where the French portion in the training dataset of XNLI is obtained by machine translation. In this task, the goal is to determine whether there is entailment, contradiction, or neutral relation from a sentence p called premise to another sentence h called hypothesis. Note that the same sample can be used twice with a reverse order between the two sentences with a different label. The dataset consists of 92 702 training samples, 2 491 validation samples and 5 010 test samples. \\
        The used model to finetune the word vectors on this task is ESIM with the same configurations and parameters described in section \ref{pawsx}.

            \begin{table}[ht]
            \centering
            \begin{tabular}{ll}
            \hline
            Embeddings  & Accuracy\\
            \hline
            w2v\_0       & 61.37    \\
            fr\_w2v\_web\_w5       & 67.71    \\
            fr\_w2v\_web\_w20    & 68.27 \\
            fr\_w2v\_fl\_w5  & 69.41   \\
            fr\_w2v\_fl\_w20   & \textbf{69.57}    \\
            \hline
            cc.fr.300   & 64.70  \\
            frWac\_500\_cbow & 63.82\\
            frWac\_200\_cbow & 63.74\\
            frWac\_700\_sg &  60.78\\
            frWiki\_1000\_cbow & 61.34\\
            \hline
            \end{tabular}
            \caption{Accuracy on the French XNLI using ESIM}
            \label{tab:xnli}
            \end{table}
            
        \textbf{Results}    We report the final accuracy of fine-tuning the various word vectors on the French XNLI dataset in Table \ref{tab:xnli}.  The results keep showing the advantages of pretrained weights and the out-performance of our CBoW Word2Vec weights over the already existing word vectors.
        
        \subsection{Noun Sense Disambiguation}
        \hspace{0.5cm}\textbf{Dataset description} This task is proposed by \citep{le2019flaubert} for the word sense disambiguation (WSD) of French that targets the nouns only. It is based on the French portion of the WSD task in \citep{inproceedings} to create the evaluation set composed of 306 sentences and 1 445 French nouns annotated with WordNet sense keys and manually verified. The training set is obtained by using the best English-French machine translation system in fairseq tool \citep{ott2019fairseq} to translate the SemCor and the WordNet Glosses Corpus to French.
        
        \textbf{Model description}   We used the same classifiers presented by \citep{vial-etal-2019-sense} that forward the output of our words vectors into a stack of 6 transformer encoder layers and predict the meaning of the word through a softmax layer at the end. We used the same parameters  as in \citep{le2019flaubert}.
        
            \begin{table}[ht]
            \centering
            \begin{tabular}{llll}
            \hline
            \multirow{0}{0pt}{Embedding}  & \multicolumn{2}{c}{Single}& \multirow{0}{0pt}{Ens}\\
            & Mean &   \hspace{0.2cm}Std \\
            \hline
            w2v\_0       & 47.85 & $\pm$1.17 & 52.87\\
            fr\_w2v\_web\_w5       & 50.76 & $\pm$1.4   & \textbf{53.77}\\
            fr\_w2v\_web\_w20    & 50.28 & $\pm$0.92 & 53.22\\
            fr\_w2v\_fl\_w5  & 50.16 & $\pm$1.41    & 53.36 \\
            fr\_w2v\_fl\_w20   & 50.65 & $\pm$1.62    & 52.46\\
            cc.fr.300   & 49.28 & $\pm$1.5  & 52.39\\
            \hline
            \end{tabular}
            \caption{F1 score (\%) on the NSD task}
            \label{tab:nsd}
            \end{table}        
        \textbf{Results}   For every word vectors, we report in Table \ref{tab:nsd} the mean and standard deviation values of F1 scores (\%) of 8 individual models. For instance, each model predicts the word with the highest probability in the softmax layer's output and the F1 score (\%) of the ensemble of models which averages out the output of the softmax layer. We observe in this task that, even though fr\_w2v\_web\_w5 slightly outperforms the other word vectors, we have a very similar F1 score (\%) for all the models. We can say that, for this task, pretrained word vectors do not improve the final score. We think that the nature of these vectors being static despite the context while the goal of the task is to investigate the different meanings of a word is the reason for the similar score between these models.   
        
\section{Conclusion}
  In this work, we contribute word embeddings learned on data crawled from the French web and other word vectors trained on a mixed dataset formed from different and diverse sources and subjects in French including common crawl and Wikipedia. These word vectors can be used as initial weights for various deep learning models which can bring important performance boost. In addition, these general domain word embeddings could be further pretrained on specific domain data such as health and legal domain in order to adapt the weights to a suitable context.\\
  Moreover, we created a benchmark of four NLP tasks to compare the quality of four produced French word embeddings and five existing ones. All the resources are presented and available to the research community via our web app.\\
  Finally, we showed that the word analogy task could not be trusted to judge on word embedding's quality. The reason behind these results on the word analogy task could be a potential future study.
\section*{Acknowledgment}
This research was supported by the ANR chair AML/HELAS (ANR-CHIA-0020-01). 

\bibliography{eacl2021}
\bibliographystyle{acl_natbib}

\end{document}